\documentclass{article}

\usepackage{cisa_preprint,times}

\usepackage[T1]{fontenc}
\usepackage[utf8]{inputenc}
\usepackage{amsmath,amssymb}
\usepackage{booktabs}
\usepackage{graphicx}
\usepackage{xcolor}
\usepackage{microtype}
\usepackage{multirow}
\usepackage{placeins}
\usepackage{url}
\usepackage{hyperref}

\newcommand{\PaperTitle}{CISA: Retrofitting Dense Attention Models with Chunk-Indexed Sparse Attention}
\newcommand{\PaperAuthors}{%
  Mohammed~Dabbah, Izhak~Golan, Oren~Tropp, Daniel~Afrimi, Vladimir~Anisimov,\\
  Tomer~Asida, Nave~Assaf, Ran~El-Yaniv, Ido~Galil, Yonatan~Geifman,\\
  Omri~Puny, Ido~Shahaf, and Akhiad~Bercovich\\[7pt]
  NVIDIA, {\ttfamily
    \{\PaperEmail{mdabbah},\,\PaperEmail{izikg},\,\PaperEmail{otropp},\,\PaperEmail{dafrimi},\,\PaperEmail{vanisimov},\,\PaperEmail{tasida},\,\PaperEmail{nassaf},\\[2pt]
    \PaperEmail{relyaniv},\,\PaperEmail{igalil},\,\PaperEmail{ygeifman},\,\PaperEmail{opuny},\,\PaperEmail{ishahaf},\,\PaperEmail{abercovich}\}@nvidia.com}%
}
\newcommand{\PaperEmail}[1]{\href{mailto:#1@nvidia.com}{\texttt{#1}}}

\newcommand{\cisa}{CISA}
\newcommand{\CISA}{Chunk-Indexed Sparse Attention}
\newcommand{\ruler}{RULER}
\newcommand{\aalcr}{AA-LCR}
\newcommand{\mrcr}{MRCR}
\newcommand{\graphwalks}{GraphWalks}
\newcommand{\longbenchv}{LongBench v2}
\newcommand{\qshare}{\textit{QShare}}
\newcommand{\topk}{top-$k$}

\newcommand{\ctxlen}{N}
\newcommand{\chunksize}{C}
\newcommand{\chunkbudget}{k_{\mathrm{chunks}}}
\newcommand{\qsharegroup}{G}
\newcommand{\indexdim}{d_I}
\newcommand{\headdim}{d}
\newcommand{\selectedset}{\mathcal{S}}

\title{\PaperTitle}
\author{\PaperAuthors}
\hypersetup{
  hidelinks,
  pdftitle={\PaperTitle},
  pdfauthor={Mohammed Dabbah, Izhak Golan, Oren Tropp, Daniel Afrimi, Vladimir Anisimov, Tomer Asida, Nave Assaf, Ran El-Yaniv, Ido Galil, Yonatan Geifman, Omri Puny, Ido Shahaf, Akhiad Bercovich}
}

\begin{document}
\maketitle

\begin{abstract}
As context grows, LLM inference becomes increasingly expensive because attention repeatedly reads a growing key-value (KV) history.
We introduce \CISA{} (\cisa), a method for retrofitting pretrained dense attention models without gradient updates.
\cisa{} selects fixed-size chunks of the KV history using compact summaries derived from the model's existing keys, avoiding token-level indexing and a separate full-length index cache.
For indexing, \cisa{} uses a single query per KV group and shares the selected chunks across the query in that group.
During prefill, we reuse chunk selections across adjacent queries to reduce indexing cost and improve KV reuse and GPU utilization.
We call this technique \qshare{}.

We evaluate \cisa{} on the NVIDIA Nemotron~3 Nano and Ultra models, as well as on Inkling-Small.
In one-layer BF16 benchmarks at batch size 32 using Ultra's per-device attention dimensions, \cisa{} accelerates core attention by $11.7\times$ for prefill queries near the end of a 1M-token prompt and $13.0\times$ during decode.
Including chunk scoring, selection, and all other measured GPU work, the corresponding speedups are $8.6\times$ and $3.0\times$.
On Ultra NVFP4, at $\chunkbudget=4{,}096$, \cisa{} keeps aggregate scores within 1.5 percentage points of dense attention on \mrcr{}, \longbenchv{}, and \graphwalks{}.
\ruler{} sweeps on Nano and Ultra BF16 up to one million tokens show scores within 2.05 percentage points of dense attention with 8--16$\times$ fewer KV reads.

When serving million-token prompts with 32 concurrent requests and 4K output tokens per request, \cisa{} achieves $2.29\times$ the end-to-end throughput of dense attention using Nemotron~3 Ultra NVFP4 on four NVIDIA B300 GPUs.
Selecting fewer chunks increases throughput to $2.61\times$ compared with dense attention; in separate \graphwalks{} evaluations, the smaller budget widens the aggregate score gap from 1.49 to 3.59 percentage points. These results expose a practical quality--speed tradeoff without requiring sparse training.

\end{abstract}

\section{Introduction}
\label{sec:introduction}

Long-context models can process code repositories, extended conversations, and million-token documents without an external retrieval step.
This capability is expensive to serve.
During dense prefill, each query attends to an increasingly long prefix, while autoregressive decode repeatedly reads the accumulated key--value (KV) cache.
IO-aware kernels reduce data movement within exact attention \citep{dao2022flashattention}, but they do not remove the growth in attended history.
At long context lengths, KV-cache data movement is a major bottleneck for latency and throughput.

Sparse attention reduces this cost by selecting a small query-dependent subset of history and applying ordinary attention only to that subset.
The challenge is to turn fewer attended tokens into faster GPU execution despite the cost of indexing and scattered KV-cache reads.
Token-level sparse attention, such as DeepSeek Sparse Attention (DSA) \citep{deepseek2025v32}, indexes and selects individual positions from the history.
Although attention accesses fewer tokens, the indexer still scans the full history, which can limit the overall speedup.
Serving also requires writing fine-grained scores and indices, issuing irregular K/V reads, and often performing an additional gather or materialization step before attention begins.
The nominal reduction in dot products can therefore overstate the realizable speedup.

A second challenge is that several strong sparse architectures are trained natively or introduced through substantial continued training \citep{yuan2025nsa,deepseek2025v32,lai2026msa}.
That path is valuable when training a new model, but it does not directly address the large installed base of available dense attention models.

\begin{figure}[b]
  \centering
  \includegraphics[width=\textwidth]{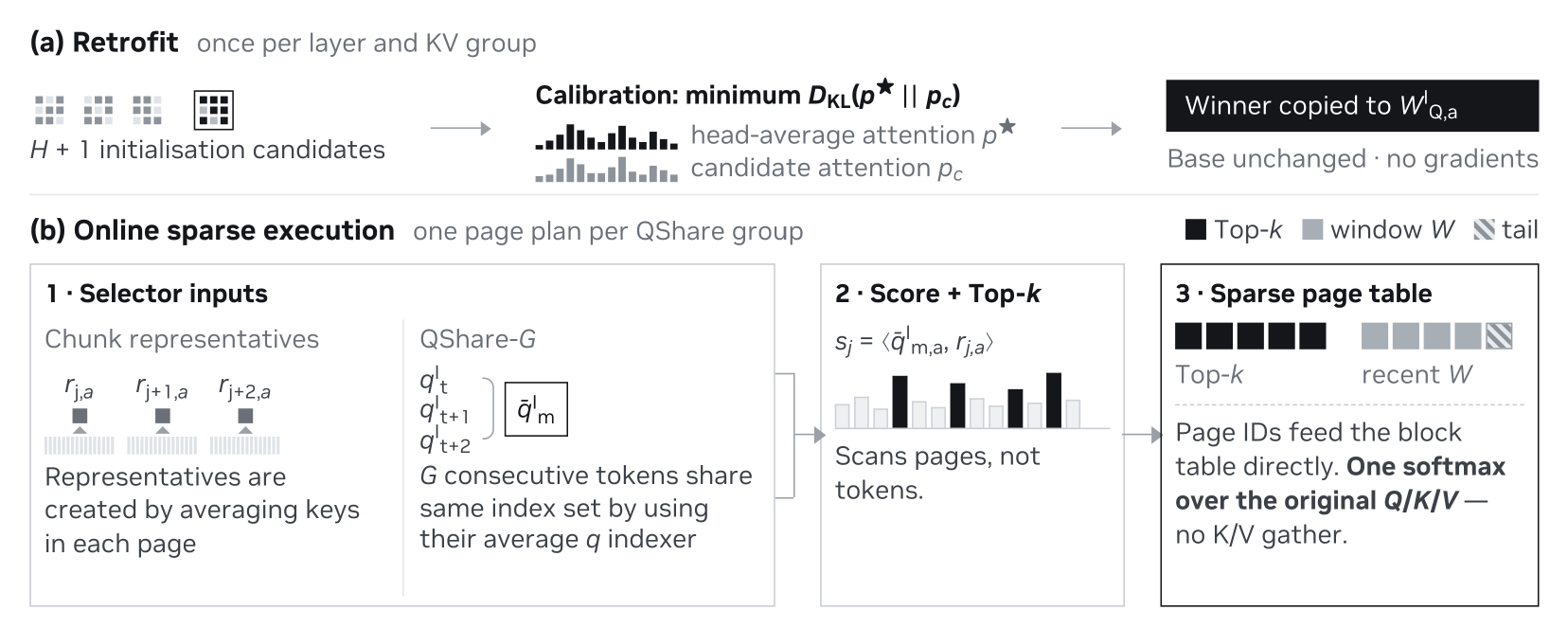}
  \caption{\cisa{} retrofit and online sparse execution. Within each KV group, \qshare{} reuses one page plan across adjacent queries; attention uses the original Q/K/V.}
  \label{fig:cisa-overview}
\end{figure}

We introduce \CISA{} (\cisa), a sparse-attention design that enables efficient long-context inference through chunk-level indexing and direct KV-cache access (Figure~\ref{fig:cisa-overview}).
\cisa{} makes three coordinated choices.
First, it indexes and selects fixed-size chunks rather than individual tokens, recalling a fixed budget of \(\chunkbudget\) historical chunks at each deployment operating point.
Second, we initialize the index-query projections from the model's existing query heads using a DSA-inspired attention-distribution matching criterion, without any gradient-based training.
Third, we share one selected chunk set across \(G\) adjacent prefill query positions. This performs indexing once per group and improves attention efficiency by reusing the selected K/V data across more query rows. We call this mechanism \qshare{}.
It is compatible with multi-token prediction (MTP).

The design is motivated by deployment topology as much as approximation quality. MiniMax Sparse Attention (MSA) selects blocks by scanning a separately projected token-level index-key stream shared across KV groups \citep{lai2026msa}; under tensor parallelism, that stream must be replicated or communicated. \cisa{} instead derives one representative per 16-token chunk from each KV group's main keys and stores it with the corresponding KV shard. Index storage and scanning scale with chunks; selected page IDs feed exact attention over original Q/K/V without a gather buffer.

We evaluate NVIDIA Nemotron~3 Nano 30B-A3B and Ultra 550B-A55B, together with Inkling-Small.
NVFP4 weights on Ultra and Inkling-Small are used for the deployment evaluations; BF16 Nano and Ultra runs provide model-scaling and context-scaling controls.
We evaluate four complementary long-context benchmarks---\mrcr{}, \longbenchv{}, \graphwalks{}, and \ruler{}---as well as \aalcr{} reasoning, SWE-bench Verified, Terminal-Bench, and GDPval \citep{vodrahalli2024michelangelo,hsieh2024ruler,bai2024longbenchv2,openai2025graphwalks,artificialanalysis2025lcr,jimenez2024swebench,merrill2026terminalbench,patwardhan2025gdpval}.
For \mrcr{}, \longbenchv{}, and \graphwalks{} on Ultra NVFP4, we apply $\chunkbudget=2{,}048$ and $\chunkbudget=4{,}096$ uniformly and report both aggregate and prompt-length-stratified quality.
We also derive an attention IO-traffic model and measure controlled, equal-work serving throughput at fixed operating points. We report theoretical component ceilings separately from measured application throughput.

This paper makes three contributions:

\begin{itemize}
  \item We introduce \cisa{}, a method for retrofitting pretrained dense attention layers with fixed-budget chunk selection, without gradient-based training.
  \item We develop a tensor-parallel-optimized sparse-attention layout with one indexer per KV group, chunk summaries from the main keys, and \qshare{} for efficient prefill.
  \item We characterize quality and generation length across two model families, chunk budgets, precisions, and context lengths up to one million tokens, linking an IO-traffic model to measured serving throughput.
\end{itemize}

\section{Method: Chunk-Indexed Sparse Attention}
\label{sec:method}

\subsection{Chunk-indexed attention for a single query head}
\label{sec:single-head}

For one query head at position $t$, let $q_t$ be the query and
$\{k_u,v_u\}_{u=1}^{t}$ its causal keys and values, with head dimension $\headdim$.
Let $K_{1:t}$ and $V_{1:t}$ stack these keys and values as rows.
Dense attention applies a softmax to query--key scores and computes a weighted
sum of values:
\begin{equation}
  o^{\mathrm{dense}}_t
  =\operatorname{softmax}\!\left(
    \frac{q_t K_{1:t}^{\top}}{\sqrt{\headdim}}
  \right)V_{1:t}.
  \label{eq:dense-attention}
\end{equation}
\cisa{} selects chunks of history, then computes attention using their
original keys and values.

We partition the token sequence into chunks $\mathcal{C}_j$ of
$\chunksize$ consecutive tokens, indexed from $j=1$.
For chunk selection, each full chunk $\mathcal{C}_j$ is represented by the
mean $r_j$ of its $\chunksize$ key vectors.
For a query at position $t$, attention always includes the tokens within the chunk containing $t$, which are at positions
$\mathcal{T}_t=\{(\lceil t/\chunksize\rceil-1)\chunksize+1,\ldots,t\}$, and the tokens $\mathcal{W}_t$ in up to $W$
immediately preceding chunks.
The eligible set $\mathcal{E}_t$ contains the indices of the full historical
chunks outside the recent window.
A separate query projection produces $q^I_t$ to score their representatives
and select a fixed budget of $\chunkbudget$ chunks:
\begin{align}
  r_j&=\frac{1}{\chunksize}\sum_{u\in\mathcal{C}_j} k_u,
  \quad s_{t,j}=\left\langle q^I_t,r_j\right\rangle,
  \qquad j\in\mathcal{E}_t
  =\left\{1,\ldots,\max\!\left(0,
    \left\lceil\frac{t}{\chunksize}\right\rceil-W-1\right)\right\},
  \label{eq:selection-score}\\
  \selectedset_t&=\operatorname{TopK}_{j\in\mathcal{E}_t}
  \bigl(s_{t,j},\chunkbudget\bigr).
  \label{eq:fixed-topk}
\end{align}
An upper bound of zero denotes an empty eligible set.
The index query serves only selection; its projection is initialized in
Section~\ref{sec:cisa-initialization}.
We use all eligible chunks if fewer than $\chunkbudget$ are available.
The set of attended token positions is
$\mathcal{U}_t=(\bigcup_{j\in\selectedset_t}\mathcal{C}_j)
\cup\mathcal{W}_t\cup\mathcal{T}_t$.
Attention uses the original Q/K/V and normalizes over this set:
\begin{equation}
  o^{\mathrm{CISA}}_t
  =\operatorname{softmax}\!\left(
    \frac{q_t K_{\mathcal{U}_t}^{\top}}{\sqrt{\headdim}}
  \right)V_{\mathcal{U}_t},
  \label{eq:sparse-attention}
\end{equation}
where $K_{\mathcal{U}_t},V_{\mathcal{U}_t}$ denote the key and value rows at
positions $\mathcal{U}_t$.
When all eligible chunks fit within the budget, this reduces to dense attention.

\subsection{One indexer per KV group}
\label{sec:kv-group-indexer}

In grouped-query attention, let $\mathcal{H}_a$ denote the set of query heads
sharing the keys and values of KV group $a$, and let $H=|\mathcal{H}_a|$.
\cisa{} uses one index query $q^I_{t,a}$ for the group:
Equations~\ref{eq:selection-score}--\ref{eq:fixed-topk} form representatives
$r_{j,a}$ from the KV group's keys and select one chunk set $\selectedset_{t,a}$, which is
shared by all $H$ query heads.
Each head still computes Equation~\ref{eq:sparse-attention} with its own query
$q_{t,h}$ and the group's original keys and values $k_{u,a},v_{u,a}$.
Thus, the heads read the same tokens but assign their own attention weights.

Representatives follow the main keys' tensor-parallel ownership and have
dimension $\indexdim=\headdim$.
For a sequence of $\ctxlen$ tokens, the representative cache stores approximately
$\ctxlen/\chunksize$ vectors per local KV group, each of dimension $\indexdim$,
avoiding a separate token-level index-key cache.

\subsection{\qshare{}: sharing chunk selection across query positions}
\label{sec:qshare}

During prefill, \qshare{} uses one chunk selection for multiple adjacent query
positions within each KV group.
For a prefix of $L_{\mathrm{prefix}}$ tokens, \qshare{}-$\qsharegroup$ partitions
the query positions into disjoint groups
$\mathcal{G}_m=\{(m-1)\qsharegroup+1,\ldots,\min(m\qsharegroup,L_{\mathrm{prefix}})\}$,
where $m=1,\ldots,\lceil L_{\mathrm{prefix}}/\qsharegroup\rceil$.
Each group contains $\qsharegroup$ consecutive query positions, except the final
group, which may be shorter.
Standard autoregressive decoding contributes one new query position per sequence,
so $\qsharegroup=1$.

Within each group $\mathcal{G}_m$ and KV group $a$, we form a shared index query
by averaging the projected index queries:
\begin{equation}
  \bar{q}^{I}_{m,a}
  =\frac{1}{|\mathcal{G}_m|}
  \sum_{t\in\mathcal{G}_m}q^{I}_{t,a}.
  \label{eq:qshare-query}
\end{equation}
We use $\bar{q}^{I}_{m,a}$ in
Equations~\ref{eq:selection-score}--\ref{eq:fixed-topk} to score the chunk
representatives and select one historical chunk set for $\mathcal{G}_m$.
Only chunk indices in $\bigcap_{t\in\mathcal{G}_m}\mathcal{E}_t$ are considered,
so every selected chunk is eligible for every position in the group.
Each head at each position then computes attention using its original query and
the original K/V from the selected chunks, recent window, and current-chunk
tail, with a position-specific causal mask.

Sharing the chunk selection amortizes representative scoring and \topk{}
selection across positions and lets more query rows reuse the selected K/V.
The resulting gains in KV reuse and GPU tile utilization are examined in
Section~\ref{sec:component-to-throughput}.

\subsection{Initializing the indexer from dense attention}
\label{sec:cisa-initialization}

We initialize each group's index-query projection from a pretrained dense
attention layer without gradient-based training.

For each converted attention layer and KV group $a$, the candidate set
$\mathcal{C}_a$ contains $H+1$ query projections: the projections of the $H$
heads in $\mathcal{H}_a$ and their rowwise average.

For each sequence $x$ and query position $t$ in the calibration set
$\mathcal{D}_{\mathrm{cal}}$, we define the indexer's target by averaging the
dense attention distributions of the $H$ query heads in KV group $a$:
\begin{equation}
  p^\star_{x,t,a}
  =
  \frac{1}{H}
  \sum_{h\in\mathcal{H}_a}
  \operatorname{softmax}
  \left(
    \frac{q_{x,t,h}K_{x,1:t,a}^{\top}}{\sqrt{\headdim}}
  \right).
  \label{eq:target-distribution}
\end{equation}

We include the averaged query projection as a candidate for approximating this
target. Because softmax is nonlinear, however, the attention distribution produced by the averaged query is not generally equal to the average of the heads' attention distributions. Each candidate projection $c\in\mathcal{C}_a$ produces a query
whose attention distribution over the same causal keys is $p^{(c)}_{x,t,a}$.
Following DSA's KL-matching objective \citep{deepseek2025v32}, we select the
candidate whose distribution most closely matches the target on the calibration set:
\begin{equation}
  c^\star_a
  =\arg\min_{c\in\mathcal{C}_a}
  \mathbb{E}_{(x,t)\sim\mathcal{D}_{\mathrm{cal}}}
  \left[
    D_{\mathrm{KL}}\!\left(
      p^\star_{x,t,a}\,\middle\|\,p^{(c)}_{x,t,a}
    \right)
  \right].
  \label{eq:init-kl}
\end{equation}

The selected projection weights are copied into the group's indexer.\footnote{For
Nemotron~3 Ultra, the averaged projection was selected for 23 of the 24
layer--KV-group pairs.}
The base-model weights remain unchanged.

\section{System Realization and Efficiency}
\label{sec:system}

\subsection{Paged execution in vLLM}

\cisa{} reads selected chunks directly from the existing KV cache.
In vLLM, this cache is organized into fixed-size memory blocks called
\emph{pages}. These pages need not be adjacent in memory; a \emph{block table}
maps their positions in the sequence to their physical storage locations
\citep{kwon2023vllm}.
We set $\chunksize=16$ to match the page size, so each selected chunk corresponds
to one cache page. \cisa{} supplies the selected physical page IDs to the existing
attention kernel, which reads the original K/V in place without gathering or
copying them into a separate buffer.

The serving implementation also caches the chunk representatives.
As full chunks become available, it updates this cache and reuses the stored
summaries during selection.
For equal index and attention head dimensions, one representative per
$\chunksize=16$ tokens requires $1/(2\chunksize)=1/32$ as many elements as the
corresponding main K+V state.
The full KV history remains stored so later queries can select different chunks.
\cisa{} reduces cache reads rather than cache capacity.

The runtime skips representative scoring and \topk{} selection when all eligible
chunks fit within the budget $\chunkbudget$.
Prefill query blocks crossing this threshold are split into a full-attention
prefix and a sparse-attention suffix.

\subsection{Attention-traffic model}

Let $\headdim$ be the attention head dimension and $\indexdim$ the representative dimension.
We analyze traffic per KV group, whose K/V loads are reused across its
assigned query heads.
For a query with history length $\ctxlen$, dense attention reads approximately $2\ctxlen\headdim$ elements from the K/V cache.
Once sparse selection is active, \cisa{} reads $2\chunkbudget\chunksize\headdim$ selected K/V elements and scans $(\ctxlen/\chunksize)\indexdim$ representative elements.
With \qshare{}-$\qsharegroup$, the representative scan is amortized across $\qsharegroup$ adjacent queries.
Ignoring the recent window, current-chunk tail, projections, metadata, and
representative construction, we express \cisa{}'s estimated memory-read volume
for a query near the end of prefill as a fraction of that required by dense attention:
\begin{equation}
  \rho_{\mathrm{end}}(\ctxlen,\chunkbudget)
  \approx
  \underbrace{\frac{\chunkbudget\chunksize}{\ctxlen}}_{\text{K/V reads}}
  +
  \underbrace{\frac{\indexdim}{2\chunksize\qsharegroup\headdim}}_{\text{representative reads}}.
  \label{eq:traffic-ratio}
\end{equation}
This per-query expression does not describe a complete causal prefill: early queries still attend to all available history until $\chunkbudget\chunksize$ tokens are available.
Let $\alpha=\chunkbudget\chunksize/\ctxlen$ denote the selected historical token
budget as a fraction of the full prefix length.
Integrating traffic over the full causal triangle gives
\begin{equation}
  \rho_{\mathrm{prefill}}
  \approx
  \underbrace{2\alpha-\alpha^2}_{\text{K/V reads}}
  +
  \underbrace{(1-\alpha^2)
  \frac{\indexdim}{2\chunksize\qsharegroup\headdim}}_{\text{representative reads}}.
  \label{eq:prefill-traffic}
\end{equation}
The inverse of either ratio is a bandwidth-only speedup estimate based on modeled memory reads, not a bound on measured kernel or end-to-end speedup.

For $\chunksize=16$, $\indexdim=\headdim$, and $\qsharegroup=4$, selecting $\chunkbudget=\frac{N}{8C}$, corresponding to 8x nominal compression of the selected remote-history span, gives a $7.53\times$ estimate near the end of prefill and a $4.13\times$ full-prefill estimate.
Selecting $\chunkbudget=\frac{N}{16C}$ (16x compression) yields $14.22\times$ and $7.76\times$, respectively.
Conventional single-token decode has $\qsharegroup=1$, reducing the corresponding per-query ceilings to $6.40\times$ and $10.67\times$.
The fixed recent window and at-most-one-chunk causal tail lower these
ceilings by a small additive term whose relative contribution decreases
with context length.

\subsection{From component savings to throughput}
\label{sec:component-to-throughput}

The traffic reductions above do not by themselves guarantee proportional time
savings: selected pages must feed well-utilized attention kernels, and scanning
the representatives must be efficient.

Ultra has 32 query heads per KV head, so $G=4$ provides $4\times32=128$
query rows per selected KV set, filling the $M=128$ FlashAttention tile used in our implementation.
Inkling-Small has four query heads per KV head: $G=4$ supplies 16 rows and
underfills the tile, whereas $G=32$ fills its $32\times4=128$ rows.
Larger sharing groups also reduce the number of separate selections, which may reduce quality.

\begin{figure}[t]
  \centering
  \includegraphics[width=0.98\textwidth]{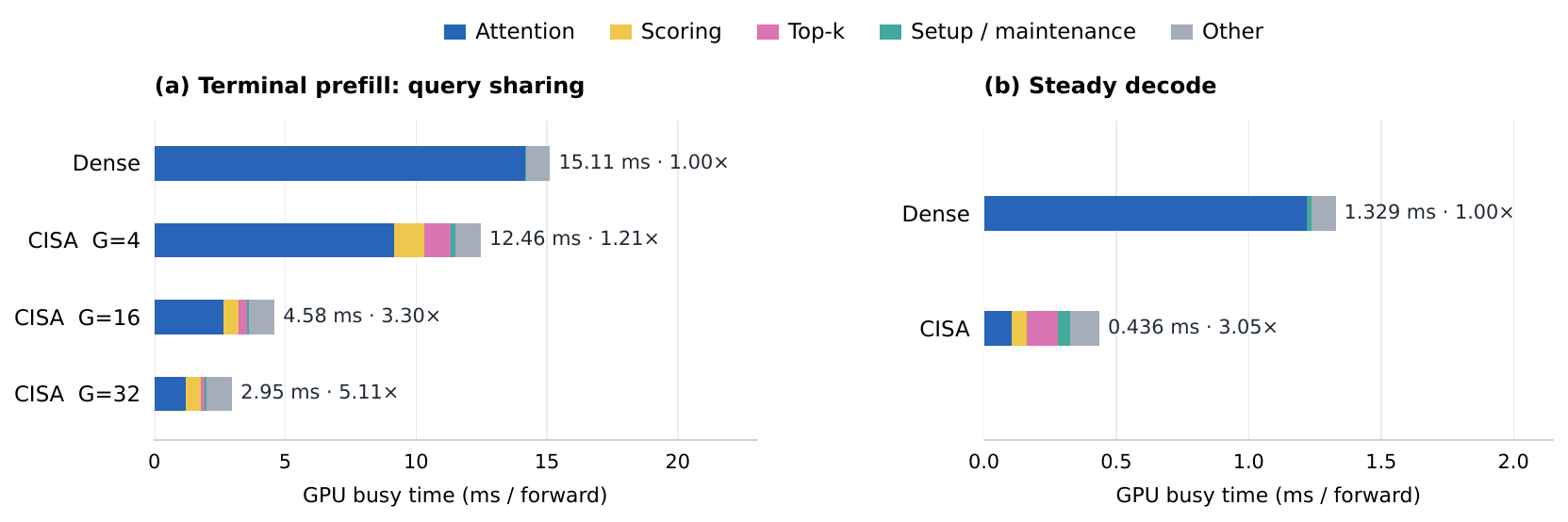}
  \input{sections/inkling_systems_caption}
  \label{fig:inkling-systems-breakdown}
\end{figure}

Figure~\ref{fig:inkling-systems-breakdown} illustrates these effects using one-layer
BF16 Inkling-Small benchmarks at 1M context, batch size 16, and $\chunkbudget=4{,}096$.
Relative to dense attention, terminal-prefill GPU-busy speedups reach
$1.21\times$ ($G=4$), $3.30\times$ ($G=16$), and $5.11\times$ ($G=32$).
Single-token decode, with effective positional sharing $G=1$, achieves $3.05\times$.

For Ultra's per-rank TP2 attention geometry at the same context and budget,
one-layer BF16 benchmarks at batch size 32 yield core-attention speedups of
$11.70\times$ for terminal prefill and $12.99\times$ for decode.
Including scoring, selection, and other measured GPU work in this one-layer benchmark, the corresponding speedups are $8.61\times$ and $2.98\times$.

Appendix~\ref{app:attention-only-speedup} provides component timings and experimental details.
These measurements isolate attention-path efficiency;
Section~\ref{sec:quality-throughput} evaluates how the gains translate into full-model
serving throughput.

\section{Evaluation}
\label{sec:evaluation}

We evaluate \cisa{} in four stages: budget-matched \ruler{} scaling, heterogeneous long-context workloads, capability preservation, and controlled quality--throughput measurements.
Unless marked otherwise, sparse runs use 16-token chunks, a 128-page recent window, and \qshare{}-4; each result states its fixed chunk budget $\chunkbudget$;
tables and figures use $k$ or $K$ as shorthand for $\chunkbudget$.
Model configurations, sampling, cohort construction, and detailed statistics appear in Appendix~\ref{app:evaluation-details}.

\subsection{Budget-matched \ruler{} scaling}

\begin{figure}[t]
  \centering
  \includegraphics[width=0.98\textwidth]{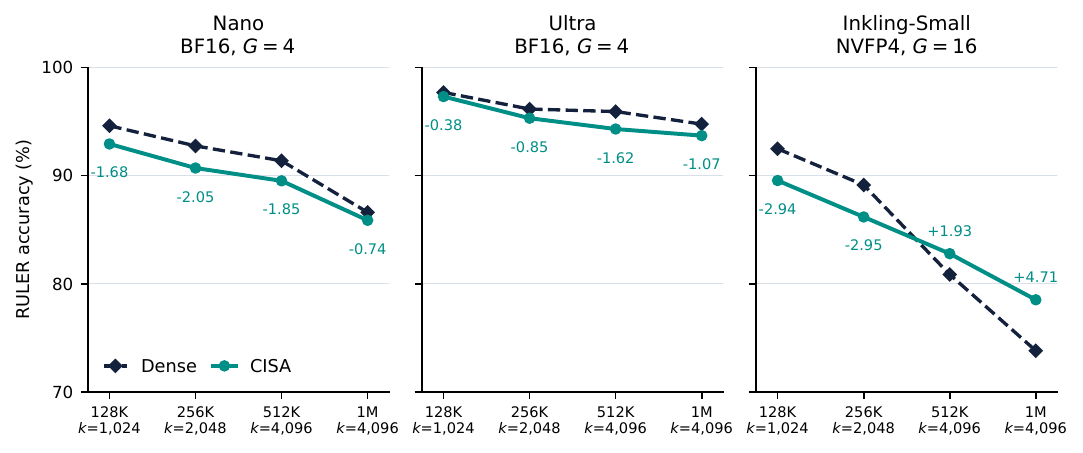}
  \caption{Budget-matched \ruler{} results for Nano and Ultra (BF16, $G=4$) and Inkling-Small (NVFP4, $G=16$).
  Each sparse run keeps the labeled budget fixed.
  Panels share the same accuracy scale; signed labels show \cisa{} minus dense accuracy in percentage points.
  Exact scores, budgets, and remote-compression factors appear in Appendix Table~\ref{tab:ruler-scaling}.}
  \label{fig:ruler-scaling}
\end{figure}

We evaluate \ruler{} scaling on Nano and Ultra with BF16 weights and on Inkling-Small with NVFP4 weights (Figure~\ref{fig:ruler-scaling}).
Nano and Ultra use $G=4$, while Inkling-Small uses $G=16$ for this \ruler{} sweep.
All three models follow the same budget schedule: $\chunkbudget=1{,}024$, $2{,}048$, $4{,}096$, and $4{,}096$ at 128K, 256K, 512K, and 1M, respectively, with each budget fixed throughout its run.
At the nominal context endpoints, these budgets select $8\times$ fewer remote historical tokens through 512K and $16\times$ fewer at 1M, before adding the recent window and current-page tail.

On Nano and Ultra, \cisa{} remains within 2.05 points of dense attention across all eight model--length comparisons, with smaller gaps at 1M than at 512K.
With $G=16$, Inkling-Small trails dense attention by 2.94 points at 128K and 2.95 points at 256K, then scores 1.93 points above dense at 512K and 4.71 points above at 1M.
Because the budget changes with context length, this sweep measures the combined operating point rather than the independent effect of either variable.
Additional \ruler{} results across recall budgets and query-sharing groups are in Appendix Table~\ref{tab:ruler-scaling}.

\subsection{Heterogeneous workloads with very long contexts}
\label{sec:long-context-workloads}

\ruler{} provides a controlled scaling sweep, but heterogeneous workloads place different demands on attention.
We therefore evaluate Ultra and Inkling-Small on \mrcr{}, \longbenchv{}, and \graphwalks{} using NVFP4 weights at $\chunkbudget\in\{2{,}048,4{,}096\}$.
We use $G=4$ for Ultra and $G=32$ for Inkling-Small, following the hardware considerations in Section~\ref{sec:component-to-throughput}.
Within each model, dense and sparse comparisons use the same tasks and repeats, with prompt-length cohorts fixed by the dense reference.
Table~\ref{tab:long-context-fixed-budget-nvfp4} summarizes aggregate accuracy and changes in mean generation length.

\begin{table}[t]
  \centering
  \caption{Ultra and Inkling-Small NVFP4 long-context accuracy and generation length at two fixed budgets.
  Ultra uses $G=4$; Inkling-Small uses $G=32$.
  Scores are percentages, and score $\Delta$ is \cisa{} minus the corresponding dense baseline in percentage points.
  Verbosity $\Delta$ is the relative change in mean generated tokens.
  The two budgets select 32,768 and 65,536 remote historical tokens, respectively, in addition to the fixed recent window and current-page tail.
  Sampling and cache settings appear in Appendix~\ref{app:sampling-settings}.}
  \label{tab:long-context-fixed-budget-nvfp4}
  \small
  \setlength{\tabcolsep}{4pt}
  \resizebox{\textwidth}{!}{\begin{tabular}{@{}lrrrrrrr@{}}
\toprule
Benchmark & Dense & \cisa{} $k=2{,}048$ & $\Delta$ & Verb. $\Delta$ & \cisa{} $k=4{,}096$ & $\Delta$ & Verb. $\Delta$ \\
\midrule
\multicolumn{8}{l}{\textbf{Ultra NVFP4} ($G=4$)} \\
\mrcr{} & 47.34 & 46.64 & -0.70 & +2.9\% & 46.96 & -0.38 & +3.1\% \\
\longbenchv{} & 61.38 & 61.38 & +0.00 & +8.0\% & 62.48 & +1.09 & +4.7\% \\
\graphwalks{} & 53.42 & 49.84 & -3.59 & +14.1\% & 51.93 & -1.49 & +10.6\% \\
\midrule
\multicolumn{8}{l}{\textbf{Inkling-Small NVFP4} ($G=32$)} \\
\mrcr{} & 41.46 & 43.33 & +1.87 & -3.8\% & 42.41 & +0.94 & -3.6\% \\
\longbenchv{} & 63.27 & 64.31 & +1.04 & +24.3\% & 63.37 & +0.10 & +3.9\% \\
\graphwalks{} & 57.33 & 56.81 & -0.52 & +128.3\% & 57.05 & -0.28 & +78.2\% \\
\bottomrule
\end{tabular}
}
\end{table}

Across both models, \mrcr{} and \longbenchv{} remain close to dense attention at both budgets.
\graphwalks{} is more sensitive to the budget on Ultra: increasing $\chunkbudget$ from 2,048 to 4,096 reduces the accuracy deficit from 3.59 to 1.49 points.
Inkling-Small's \graphwalks{} scores remain closer to dense, with deficits of 0.52 and 0.28 points at the smaller and larger budgets, respectively.

The breakdown by prompt length shows where Ultra benefits from the larger budget (Appendix Figure~\ref{fig:long-context-budget-by-length-full} and Table~\ref{tab:fixed-budget-by-length-nvfp4}).
In the 256--512K \graphwalks{} cohort, increasing $\chunkbudget$ from 2,048 to 4,096 narrows the gap from 7.24 to 1.79 points.
At one million tokens and above, however, dense \graphwalks{} accuracy falls to 7.08\%, and both sparse settings remain below it.
\mrcr{} tracks dense attention across the length range.
\longbenchv{} remains within five points through 512K, while its longer cohorts have wide uncertainty because they contain only 12, 7, and 33 prompts.
Appendix Figure~\ref{fig:inkling-long-context-budget-by-length} reports the corresponding breakdown for Inkling-Small.

On Inkling-Small's \graphwalks{} evaluation, \cisa{} generates 128.3\% more tokens on average than dense attention at $\chunkbudget=2{,}048$ and 78.2\% more at $\chunkbudget=4{,}096$.
Inspection reveals longer reasoning, accompanied by some repetitive generations that exhaust the output budget.

\subsection{Reasoning, agentic, and professional capability}
Beyond long-context retrieval, we evaluate whether \cisa{} preserves the model's reasoning, agentic, and professional-task capabilities.
Table~\ref{tab:capability-quality-main} reports the Nano control and the primary Ultra NVFP4 capability results, including generation-length changes; the Ultra BF16 controls and full matrix remain in Appendix Table~\ref{tab:capability-quality-full}.

\begin{table}[t]
  \centering
  \caption{Capability preservation for Nano BF16 and Ultra NVFP4 at their target budgets.
  Scores use each benchmark's native percentage metric except GDPval, reported in Elo; score $\Delta$ is in percentage points or Elo, respectively, and verbosity $\Delta$ is the relative change in mean generated tokens.}
  \label{tab:capability-quality-main}
  \small
  \setlength{\tabcolsep}{3pt}
  \begin{tabular}{@{}llrrrrr@{}}
\toprule
Model & Benchmark & $k$ & Dense & \cisa{} & $\Delta$ & Verbosity $\Delta$ \\
\midrule
Nano BF16 & \aalcr{} & 1,024 & 34.94 & 34.25 & -0.69 & -9.1\% \\
\midrule
Ultra NVFP4 & \aalcr{} & 1,024 & 65.00 & 65.88 & +0.88 & +1.8\% \\
 & SWE-bench Verified & 2,048 & 69.20 & 69.93 & +0.73 & -0.2\% \\
 & Terminal-Bench 2.0 & 2,048 & 48.03 & 49.44 & +1.40 & -0.7\% \\
 & Terminal-Bench Hard & 1,024 & 35.94 & 35.42 & -0.52 & +2.8\% \\
 & GDPval & 2,048 & 1395.59 & 1408.71 & +13.12 & -2.8\% \\
\bottomrule
\end{tabular}

\end{table}

Across the percentage-scored Ultra NVFP4 evaluations in Table~\ref{tab:capability-quality-main}, \cisa{} changes the point estimates by $-0.52$ to $+1.40$ points; the Terminal-Bench rows use context summarization and therefore do not establish unchanged compaction behavior.
On GDPval the failure-adjusted Elo point estimates are  1395.59 for dense attention and 1408.71 for \cisa{}, a difference of $+13.12$.
Appendix~\ref{app:capability-matrix} details the adjustment and the matched-rollout comparison.

\subsection{Quality, Throughput, and Output Length}
\label{sec:quality-throughput}

We evaluate how recall budget and query sharing jointly affect quality, measured
throughput, and mean output length on Ultra and Inkling-Small.
We define quality as the equal-weight mean score across \ruler{} 1M, MRCR200,
\graphwalks{}-200, and \aalcr{}.
The base profile selects 4,096 historical chunks for the first three benchmarks
and 1,024 for \aalcr{}. The reduced profile halves these budgets.
We vary $G$ across profiles.
Both models use the same tasks and reference answers, and within each model quality and output length are measured on the same generations.

We measure controlled fixed-output total token throughput on MRCR200, the 200 longest prompts in our
\mrcr{} corpus, with contexts around one million tokens.
Every configuration serves the same prompts, and 
each request generates exactly 4,096 tokens, fixing the input and output workload
across configurations.
We report speedup $S$ relative to the same model's dense baseline.

In the quality evaluations, some \cisa{} configurations generate more output
tokens per response on average than dense attention.
In deployment, longer responses could offset part of the benefit of faster token generation.
For each benchmark, we divide \cisa{}'s mean output token count, including
reasoning tokens, by that of the dense baseline.
We average the four ratios with equal weight to obtain $R$, then define
the verbosity-adjusted speedup proxy $S_{\mathrm{adj}}=S/R$.
This simple adjustment assumes that workload cost scales linearly with output length. It does not model fixed prefix cost, nonlinear batching effects, or differences between the quality and throughput workloads.

\begin{figure}[t]
  \centering
  \includegraphics[width=0.98\textwidth]{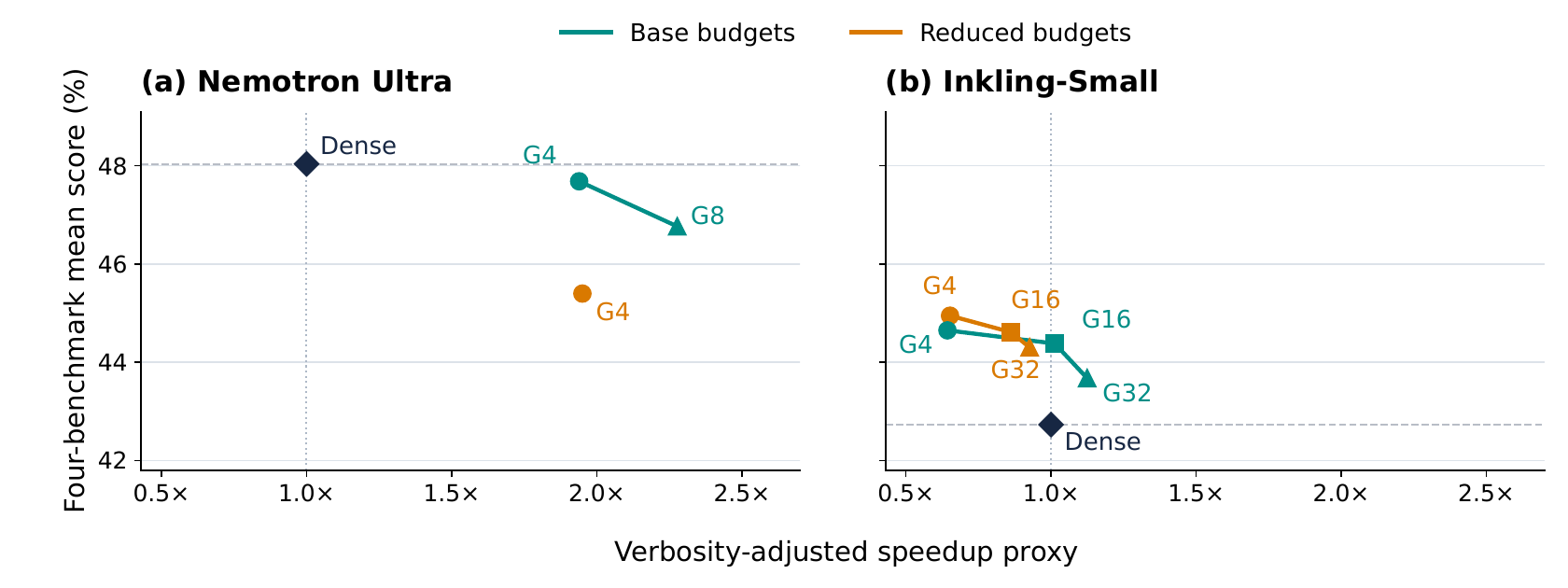}
  \caption{Quality--cost operating points for (a) Nemotron Ultra and (b) Inkling-Small, both NVFP4.
  Quality is the equal-weight mean score across \ruler{} 1M, MRCR200, \graphwalks{}-200, and \aalcr{}.
  The horizontal axis divides controlled, fixed-output total token throughout speedup by the mean benchmark-relative output-length ratio. It is an estimate that accounts for differences in output length, not a direct measurement of natural completion throughput.
  Reduced profiles halve the base recall budgets.
  Lines connect point estimates as $G$ increases within each profile.
  Exact coordinates and measurement details appear in Appendix Table~\ref{tab:quality-throughput-adjusted} and Appendix~\ref{app:adjusted-operating-points}.}
  \label{fig:quality-throughput}
\end{figure}

Figure~\ref{fig:quality-throughput} shows the resulting trade-offs.
For Ultra, increasing sharing from $G=4$ to $G=8$ at the base budgets gives
a higher mean score point estimate and a higher adjusted speedup proxy than halving
the budgets at $G=4$.
For Inkling-Small at $G=32$, the base profile's measured $1.54\times$ speedup corresponds to an adjusted proxy of $1.12\times$, while its mean score remains above the dense reference.
The reduced profile achieves a larger measured speedup of $1.64\times$, but its
higher mean output token count lower the adjusted proxy to $0.93\times$.
Together, these results illustrate why recall budget, query sharing, and output length should be considered jointly.
Appendix Figure~\ref{fig:quality-throughput-unadjusted} reports the unadjusted
speedups, and Appendix~\ref{app:adjusted-operating-points} gives the evaluation
subsets and measurement details.

\section{Related Work}
\label{sec:related}

FlashAttention tiles exact attention to reduce IO \citep{dao2022flashattention}, while PagedAttention and vLLM manage non-contiguous KV pages and continuous batching \citep{kwon2023vllm}.
\cisa{} retains paged KV management but supplies a query-dependent sparse block table.

Quest scores KV pages using coordinatewise key bounds and produces a separate plan per query head \citep{tang2024quest}.
This design is natural for MAH, where each query head has its own KV states.
Under GQA, however, separate plans for each query head can reduce reuse among query heads that share the same KV head; \cisa{} instead creates one plan per KV group, and \qshare{} reuses that plan across query positions.
On Nano, at a 1M-token context, the hybrid 
Quest-bound $+$ \cisa{} index control scores 84.95 at $G=1$ and 84.41 at $G=4$ compared with 85.87 for the budget-aligned \cisa{} reference (Appendix Table~\ref{tab:quest-selector-controls}).

Longformer and BigBird train with fixed sparse structure \citep{beltagy2020longformer,zaheer2020bigbird}; NSA, DSA, and MSA learn selectors with GPU-aware sparse execution \citep{yuan2025nsa,deepseek2025v32,lai2026msa}.
\cisa{} does not claim block selection, per-group indexing, or training-free sparsification in isolation; it combines fixed-budget recall, a dense-initialized indexer, TP-local main-key summaries, and cross-position reuse in one paged-attention path.

\section{Limitations}
\label{sec:limitations}

Initialization requires no gradient updates but uses calibration activations and candidate selection; it is not data-free.
\cisa{} reduces attention reads but retains the main K/V state for future selection; it is not KV-cache eviction or capacity compression.
Quality also remains workload- and budget-dependent: at Ultra NVFP4, $\chunkbudget=4{,}096$ leaves the $\geq1$M \graphwalks{} cohort 4.62 points below dense attention.
We test two model families, one serving stack, and a limited budget grid.

The controlled serving workload combines prefill and decode, while the analytical ceilings omit selection, metadata.
Finally, conventional autoregressive decode exposes one query at a time, so \qshare{} currently benefits prefill only.
Extending selection reuse to MTP \citep{gloeckle2024mtp} or speculative decoding remains future work.

\section{Conclusion}

Retrofitting dense attention is both an approximation problem and a systems problem.
\cisa{} addresses them together through chunk-level recall, KV-group-local indexing, and shared prefill selection, without requiring gradient updates.
Results on Nemotron and Inkling-Small show that this design can keep many benchmark scores close to dense attention  while improving fixed-output throughput.
The trade-off remains workload-dependent: longer generations can reduce the serving benefit even though benchmark scores remain close. These results support considering
recall budget and query sharing jointly during deployment tuning. Future work includes
learned indexers, adaptive budgets, and decode-side selection reuse.

\appendix
\section{Additional Evaluation Details}
\label{app:evaluation-details}

This appendix specifies the evaluation settings and cohorts, then reports the
length-stratified results and capability controls underlying Section~\ref{sec:evaluation}.

\subsection{Sampling and execution settings}
\label{app:sampling-settings}

\paragraph{Model configurations.}
Nano and Ultra are hybrid Mamba--Transformer MoE models.
Inkling-Small combines global and sliding-window attention with MoE and short-convolution components.
Their configurations are summarized below; query and KV head counts refer to global attention, with head dimension 128 throughout.

\begin{center}
  \small
  \setlength{\tabcolsep}{4pt}
  \begin{tabular}{@{}lrrrrr@{}}
    \toprule
    Model & \shortstack{Total\\blocks/layers} & \shortstack{Global-attention\\layers} & \shortstack{Hidden\\dimension} & \shortstack{Query\\heads} & \shortstack{KV\\heads} \\
    \midrule
    Nano & 52 & 6 & 2,688 & 32 & 2 \\
    Ultra & 108 & 12 & 8,192 & 64 & 2 \\
    Inkling-Small & 42 & 7 & 4,096 & 32 & 8 \\
    \bottomrule
  \end{tabular}
\end{center}

Inkling-Small's remaining 35 layers use sliding-window attention; \cisa{} replaces only its seven global-attention layers.
Across all three models, \cisa{} preserves the base-model weights and adds initialized index-query projections.

\paragraph{Sampling.}
For Ultra and Inkling-Small, \mrcr{}, \longbenchv{}, and \graphwalks{} evaluations use top-$p=1$, four repeats per task, and a 65,536-token output cap.
\ruler{} uses temperature $10^{-5}$ and top-$p=0.99$, with 100 examples per subtask across 13 subtasks.

The capability evaluations use temperature 1 and the following task counts, repeats, and top-$p$ settings.

\begin{center}
  \small
  \setlength{\tabcolsep}{4pt}
  \begin{tabular}{@{}lrrl@{}}
    \toprule
    Benchmark & Tasks & Repeats per task & Top-$p$ \\
    \midrule
    \aalcr{} & 100 & 16 & 0.95 \\
    SWE-bench Verified & 500 & 3 & 0.95 \\
    Terminal-Bench~2.0 & 89 & 8 & 0.95 (BF16); 1 (NVFP4) \\
    Terminal-Bench Hard & 48 & 8 & 0.95 \\
    GDPval & 220 & 2 & 0.95 \\
    \bottomrule
  \end{tabular}
\end{center}

The separate \aalcr{} operating-point experiments retain 100 tasks and 16 repeats, using top-$p=1$ for Ultra and top-$p=0.95$ for Inkling-Small.
GDPval's treatment of incomplete rollouts is specified in Appendix~\ref{app:capability-matrix}.

\paragraph{Execution settings.}
The Ultra NVFP4 \graphwalks{} runs in Table~\ref{tab:long-context-fixed-budget-nvfp4} use BF16 KV and Mamba caches, with FP16 Mamba SSM state and the FlashInfer Mamba backend.

\subsection{Fixed budgets by input length}
\label{app:fixed-budget-results}

We report results by input length to examine how quality preservation varies with context length at a fixed chunk budget.
Cohorts use each model's dense-reference prompt-token counts and remain fixed across dense and \cisa{} runs.
Ultra uses the same cohorts for BF16 and NVFP4; Inkling-Small uses its own token counts, so cohort membership can differ between models.

\begin{figure}[p]
  \centering
  \includegraphics[width=0.98\textwidth]{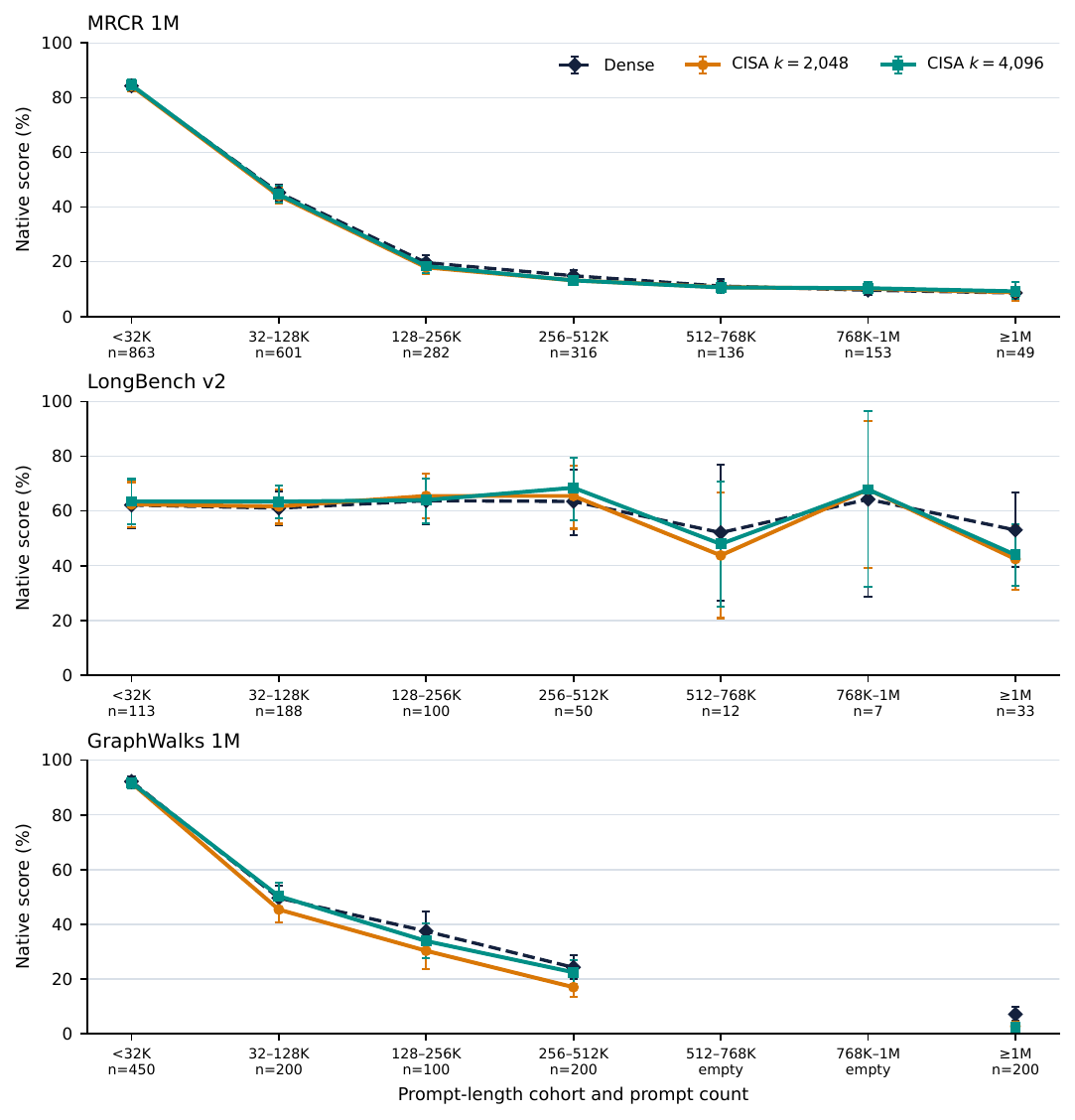}
  \caption{Ultra NVFP4 scores by dense-reference prompt length, comparing dense attention with \cisa{} at $\chunkbudget\in\{2{,}048,4{,}096\}$ and $G=4$.
  Labels give the number of distinct prompts.
  Error bars show 95\% task-bootstrap confidence intervals, retaining all four repeats of each sampled task.
  Table~\ref{tab:fixed-budget-by-length-nvfp4} reports exact scores and cohort sizes.}
  \label{fig:long-context-budget-by-length-full}
\end{figure}

\begin{figure}[p]
  \centering
  \includegraphics[width=0.98\textwidth]{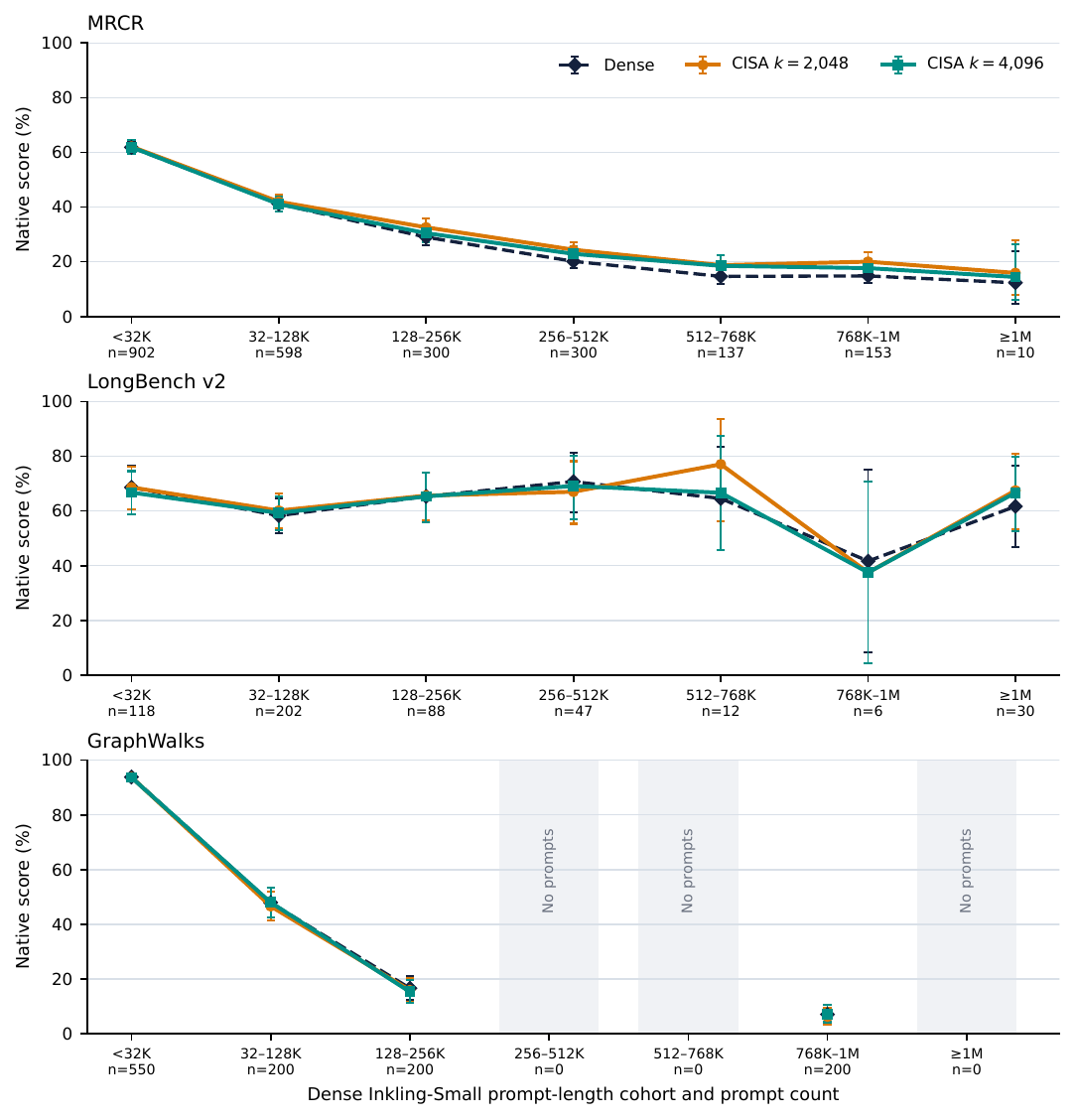}
  \caption{Inkling-Small NVFP4 scores by dense-reference prompt length, comparing dense attention with \cisa{} at $\chunkbudget\in\{2{,}048,4{,}096\}$ and $G=32$.
  Labels give the number of distinct prompts; shaded buckets contain no prompts.
  Error bars show pointwise 95\% confidence intervals from 20,000 task-bootstrap resamples, retaining all four repeats of each sampled task.}
  \label{fig:inkling-long-context-budget-by-length}
\end{figure}

Figures~\ref{fig:long-context-budget-by-length-full} and~\ref{fig:inkling-long-context-budget-by-length} show NVFP4 results for Ultra and Inkling-Small, respectively.
Each \cisa{} configuration keeps its budget of $\chunkbudget=2{,}048$ or $4{,}096$ fixed throughout the run.
Tables~\ref{tab:fixed-budget-by-length-bf16} and~\ref{tab:fixed-budget-by-length-nvfp4} provide Ultra's BF16 and NVFP4 scores and cohort sizes.

\FloatBarrier

\begin{table}[!htbp]
  \centering
  \caption{Ultra BF16 scores by dense-reference prompt length (\%).
  Both \cisa{} configurations use 16-token chunks, a 128-page recent window, and $G=4$.
  $\Delta$ is \cisa{} minus dense in percentage points.
  Empty \mrcr{} final answers are scored as zero: one for dense attention and two for $\chunkbudget=4{,}096$.}
  \label{tab:fixed-budget-by-length-bf16}
  \scriptsize
  \setlength{\tabcolsep}{3pt}
  \resizebox{\textwidth}{!}{\begin{tabular}{@{}llrrrrrrr@{}}
\toprule
Benchmark & Input length & Prompts & Rollouts & Dense & $k=2{,}048$ & $\Delta$ & $k=4{,}096$ & $\Delta$ \\
\midrule
\multirow{7}{*}{\mrcr{}} & $<32$K & 863 & 3,452 & 84.67 & 84.85 & +0.18 & 84.73 & +0.06 \\
 & 32--128K & 601 & 2,404 & 46.58 & 46.47 & -0.11 & 46.07 & -0.51 \\
 & 128--256K & 282 & 1,128 & 18.36 & 17.78 & -0.58 & 18.34 & -0.02 \\
 & 256--512K & 316 & 1,264 & 10.36 & 7.64 & -2.72 & 9.50 & -0.86 \\
 & 512--768K & 136 & 544 & 5.66 & 4.96 & -0.70 & 5.20 & -0.46 \\
 & 768K--1M & 153 & 612 & 4.76 & 4.07 & -0.69 & 4.69 & -0.07 \\
 & $\geq 1$M & 49 & 196 & 5.02 & 4.43 & -0.59 & 4.74 & -0.27 \\
\midrule
\multirow{7}{*}{\longbenchv{}} & $<32$K & 113 & 452 & 61.95 & 62.17 & +0.22 & 62.17 & +0.22 \\
 & 32--128K & 188 & 752 & 62.90 & 62.23 & -0.66 & 63.43 & +0.53 \\
 & 128--256K & 100 & 400 & 65.50 & 68.50 & +3.00 & 66.50 & +1.00 \\
 & 256--512K & 50 & 200 & 63.50 & 64.50 & +1.00 & 68.00 & +4.50 \\
 & 512--768K & 12 & 48 & 41.67 & 54.17 & +12.50 & 50.00 & +8.33 \\
 & 768K--1M & 7 & 28 & 53.57 & 57.14 & +3.57 & 64.29 & +10.71 \\
 & $\geq 1$M & 33 & 132 & 53.79 & 50.76 & -3.03 & 61.36 & +7.58 \\
\midrule
\multirow{7}{*}{\graphwalks{}} & $<32$K & 450 & 1,800 & 93.82 & 93.37 & -0.45 & 92.99 & -0.83 \\
 & 32--128K & 200 & 800 & 53.39 & 50.92 & -2.47 & 54.09 & +0.69 \\
 & 128--256K & 100 & 400 & 38.82 & 34.74 & -4.07 & 35.11 & -3.70 \\
 & 256--512K & 200 & 800 & 23.98 & 18.87 & -5.11 & 20.84 & -3.15 \\
 & 512--768K & 0 & 0 & -- & -- & -- & -- & -- \\
 & 768K--1M & 0 & 0 & -- & -- & -- & -- & -- \\
 & $\geq 1$M & 200 & 800 & 7.36 & 4.23 & -3.12 & 5.35 & -2.01 \\
\bottomrule
\end{tabular}
}
\end{table}

\begin{table}[!htbp]
  \centering
  \caption{Ultra NVFP4 scores by dense-reference prompt length (\%), with BF16 KV state.
  Both \cisa{} configurations use 16-token chunks, a 128-page recent window, and $G=4$.
  $\Delta$ is \cisa{} minus dense in percentage points.}
  \label{tab:fixed-budget-by-length-nvfp4}
  \scriptsize
  \setlength{\tabcolsep}{3pt}
  \resizebox{\textwidth}{!}{\begin{tabular}{@{}llrrrrrrr@{}}
\toprule
Benchmark & Input length & Prompts & Rollouts & Dense & $k=2{,}048$ & $\Delta$ & $k=4{,}096$ & $\Delta$ \\
\midrule
\multirow{7}{*}{\mrcr{}} & $<32$K & 863 & 3,452 & 84.32 & 84.30 & -0.02 & 84.64 & +0.32 \\
 & 32--128K & 601 & 2,404 & 45.28 & 44.13 & -1.15 & 44.64 & -0.64 \\
 & 128--256K & 282 & 1,128 & 19.65 & 17.93 & -1.72 & 18.39 & -1.26 \\
 & 256--512K & 316 & 1,264 & 14.87 & 13.17 & -1.70 & 13.17 & -1.70 \\
 & 512--768K & 136 & 544 & 11.09 & 10.70 & -0.39 & 10.60 & -0.50 \\
 & 768K--1M & 153 & 612 & 9.59 & 10.13 & +0.54 & 10.33 & +0.74 \\
 & $\geq 1$M & 49 & 196 & 8.59 & 8.80 & +0.21 & 9.18 & +0.59 \\
\midrule
\multirow{7}{*}{\longbenchv{}} & $<32$K & 113 & 452 & 62.17 & 62.39 & +0.22 & 63.50 & +1.33 \\
 & 32--128K & 188 & 752 & 61.04 & 61.70 & +0.66 & 63.43 & +2.39 \\
 & 128--256K & 100 & 400 & 63.75 & 65.50 & +1.75 & 64.00 & +0.25 \\
 & 256--512K & 50 & 200 & 63.50 & 65.50 & +2.00 & 68.50 & +5.00 \\
 & 512--768K & 12 & 48 & 52.08 & 43.75 & -8.33 & 47.92 & -4.17 \\
 & 768K--1M & 7 & 28 & 64.29 & 67.86 & +3.57 & 67.86 & +3.57 \\
 & $\geq 1$M & 33 & 132 & 53.03 & 42.42 & -10.61 & 43.94 & -9.09 \\
\midrule
\multirow{7}{*}{\graphwalks{}} & $<32$K & 450 & 1,800 & 92.23 & 91.70 & -0.53 & 91.76 & -0.47 \\
 & 32--128K & 200 & 800 & 49.60 & 45.39 & -4.21 & 50.31 & +0.71 \\
 & 128--256K & 100 & 400 & 37.48 & 30.31 & -7.18 & 33.87 & -3.61 \\
 & 256--512K & 200 & 800 & 24.22 & 16.99 & -7.24 & 22.43 & -1.79 \\
 & 512--768K & 0 & 0 & -- & -- & -- & -- & -- \\
 & 768K--1M & 0 & 0 & -- & -- & -- & -- & -- \\
 & $\geq 1$M & 200 & 800 & 7.08 & 2.70 & -4.38 & 2.46 & -4.62 \\
\bottomrule
\end{tabular}
}
\end{table}

For Ultra, the three longest \longbenchv{} cohorts contain only 12, 7, and 33 prompts and should be interpreted cautiously.
Its two intermediate \graphwalks{} cohorts between 512K and 1M are empty.

\FloatBarrier
\subsection{RULER controls}

Table~\ref{tab:ruler-scaling} reports the scores underlying Figure~\ref{fig:ruler-scaling} and additional budget and sharing controls.
Inkling-Small's dense and six sparse 128K runs use identical questions and reference answers across three budgets at $G=16$ and $G=32$.

\begin{table}[!htbp]
  \centering
  \caption{\ruler{} accuracy across context lengths, recall budgets, and query-sharing groups.
  (a) Nano and Ultra BF16 controls at $G=4$; $\Delta$ is \cisa{} minus dense.
  (b) Inkling-Small NVFP4 results at $G=16$ and $G=32$; $\Delta_{16-32}$ is the score at $G=16$ minus that at $G=32$.
  $K$ counts selected remote 16-token chunks; remote compression excludes the recent window and current tail.
  Scores are percentages and differences are percentage points, calculated before rounding.
  Daggers identify the $G=16$ settings plotted in Figure~\ref{fig:ruler-scaling}; dashes denote unavailable results.
  The Nano 1M Quest-bound controls use the same serving build, budget, window, and page size at $G=1$ and $4$; the \cisa{} reference uses a different build.}
  \label{tab:ruler-scaling}
  \label{tab:quest-selector-controls}
  \small
  \setlength{\tabcolsep}{5pt}
  \begin{tabular}{@{}llrrrrr@{}}
\toprule
\multicolumn{7}{c}{\textbf{(a) Nano and Ultra BF16: matched budgets, $G=4$}} \\
\midrule
Model & Context & $k$ (chunks) & Remote comp. & Dense & \cisa{} & $\Delta$ \\
\midrule
Nano & 128K & 1,024 & 8$\times$ & 94.61 & 92.93 & -1.68 \\
Nano & 256K & 2,048 & 8$\times$ & 92.75 & 90.70 & -2.05 \\
Nano & 512K & 4,096 & 8$\times$ & 91.37 & 89.52 & -1.85 \\
Nano & 1M & 4,096 & 16$\times$ & 86.60 & 85.87 & -0.74 \\
\multicolumn{4}{@{}l}{\quad Quest bound + \cisa{} index, Nano 1M} & \multicolumn{3}{l@{}}{$G=1$: 84.95; $G=4$: 84.41} \\
\midrule
Ultra & 128K & 1,024 & 8$\times$ & 97.69 & 97.31 & -0.38 \\
Ultra & 256K & 2,048 & 8$\times$ & 96.15 & 95.31 & -0.85 \\
Ultra & 512K & 4,096 & 8$\times$ & 95.92 & 94.31 & -1.62 \\
Ultra & 1M & 4,096 & 16$\times$ & 94.76 & 93.70 & -1.07 \\
\bottomrule
\end{tabular}

\par\vspace{6pt}
\begin{tabular}{@{}lrrrrrr@{}}
\toprule
\multicolumn{7}{c}{\textbf{(b) Inkling-Small NVFP4: budget and sharing controls}} \\
\midrule
Context & $K$ (chunks) & Remote comp. & Dense & $G=16$ & $G=32$ & $\Delta_{16-32}$ \\
\midrule
\multirow{3}{*}{128K} & 1,024 & 8$\times$ & \multirow{3}{*}{92.48} & 89.55$^{\dagger}$ & 86.83 & +2.71 \\
 & 2,048 & 4$\times$ &  & 91.71 & 90.16 & +1.55 \\
 & 4,096 & 2$\times$ &  & 92.99 & 92.22 & +0.77 \\
\midrule
\multirow{2}{*}{256K} & 2,048 & 8$\times$ & \multirow{2}{*}{89.13} & 86.18$^{\dagger}$ & 86.53 & -0.35 \\
 & 4,096 & 4$\times$ &  & \textemdash & 87.71 & \textemdash \\
\midrule
\multirow{2}{*}{512K} & 2,048 & 16$\times$ & \multirow{2}{*}{80.85} & \textemdash & 81.27 & \textemdash \\
 & 4,096 & 8$\times$ &  & 82.78$^{\dagger}$ & 81.33 & +1.45 \\
\midrule
\multirow{2}{*}{1M} & 2,048 & 32$\times$ & \multirow{2}{*}{73.81} & 76.94 & 74.53 & +2.41 \\
 & 4,096 & 16$\times$ &  & 78.52$^{\dagger}$ & 76.03 & +2.49 \\
\bottomrule
\end{tabular}

\end{table}

\subsection{Full capability results}
\label{app:capability-matrix}

Nano and Ultra use $\chunkbudget=1{,}024$ for \aalcr{} and Terminal-Bench Hard, whose dense-reference 99th-percentile input lengths are approximately 123K and 75K.
SWE-bench Verified, Terminal-Bench~2.0, and GDPval use $\chunkbudget=2{,}048$.
Inkling-Small's separate \aalcr{} budget and sharing sweep is included in Appendix~\ref{app:adjusted-operating-points}.

\begin{table}[!htbp]
  \centering
  \caption{Nano and Ultra capability results at fixed chunk budgets.
  Scores are percentages except GDPval, for which we report failure-adjusted Elo.
  Verbosity $\Delta$ is $100(\bar L_{\cisa}/\bar L_{\mathrm{dense}}-1)$, where $\bar L$ is mean generated tokens per rollout on the matched cohort.
  For multi-turn agents, rollout length sums generated tokens across all model calls.
  \textsuperscript{2} Unavailable results reflect incomplete evaluations or unrecoverable token counts for the matched cohort.
  \textsuperscript{3} GDPval includes all 220 tasks and two rollouts per task under the failure-adjustment rule below.}
  \label{tab:capability-quality-full}
  \small
  \setlength{\tabcolsep}{4pt}
  \resizebox{\textwidth}{!}{\begin{tabular}{@{}llllrrrrr@{}}
\toprule
Scale & Precision & Benchmark & Metric & $k$ & Dense & \cisa{} & $\Delta$ & Verbosity $\Delta$ \\
\midrule
Nano & BF16 & \aalcr{} & pass@1 (\%) & 1,024 & 34.94 & 34.25 & -0.69 & -9.1\% \\
\midrule
Ultra & BF16 & \aalcr{} & pass@1 (\%) & 1,024 & 65.13 & 65.38 & +0.25 & +5.3\% \\
Ultra & NVFP4 & \aalcr{} & pass@1 (\%) & 1,024 & 65.00 & 65.88 & +0.88 & +1.8\% \\
\midrule
Ultra & BF16 & SWE-bench Verified & resolved (\%) & 2,048 & 70.00 & 69.07 & -0.93 & -4.2\% \\
Ultra & NVFP4 & SWE-bench Verified & resolved (\%) & 2,048 & 69.20 & 69.93 & +0.73 & -0.2\% \\
\midrule
Ultra & BF16 & Terminal-Bench 2.0 & pass@1 (\%) & 2,048 & 49.86 & 49.30 & -0.56 & -19.1\% \\
Ultra & NVFP4 & Terminal-Bench 2.0 & pass@1 (\%) & 2,048 & 48.03 & 49.44 & +1.40 & -0.7\% \\
\midrule
Ultra & BF16 & Terminal-Bench Hard & pass@1 (\%) & 1,024 & 37.24 & 36.98 & -0.26 & -6.9\% \\
Ultra & NVFP4 & Terminal-Bench Hard & pass@1 (\%) & 1,024 & 35.94 & 35.42 & -0.52 & +2.8\% \\
\midrule
Ultra & BF16 & GDPval\textsuperscript{3} & Adj. Elo & 2,048 & 1422.62 & 1414.72 & -7.90 & --\textsuperscript{2} \\
Ultra & NVFP4 & GDPval\textsuperscript{3} & Adj. Elo & 2,048 & 1395.59 & 1408.71 & +13.12 & -2.8\% \\
\bottomrule
\end{tabular}
}
\end{table}

\paragraph{Terminal-Bench and cache precision.}
The Terminal-Bench runs use context summarization; their scores do not establish unchanged compaction frequency or fidelity.
All NVFP4 rows use BF16 KV state.
Ultra's BF16 dense Terminal-Bench references use FP8 KV caches, so these comparisons change both sparsity and KV precision.

\paragraph{GDPval scoring.}
Each rollout is compared with Kimi-K2.5-Thinking (vadams r2) using four Gemini~3.1 Pro Preview judgments with alternating response positions.
A rollout missing any of the four valid judgments at the reporting cutoff is assigned four losses.
Failure-adjusted Elo therefore includes all 440 rollouts and 1,760 comparison slots.
Complete judgments are available for 427/440 dense and 418/440 \cisa{} rollouts in BF16, and 420/440 dense and 430/440 \cisa{} rollouts in NVFP4.
Restricting the calculation to the 420 NVFP4 rollouts with complete judgments for both methods gives Elo 1419.57 for dense and 1421.94 for \cisa{} ($+2.38$, calculated before rounding).
NVFP4 verbosity counts generated tokens across all model calls in one retained trajectory per rollout for all 440 rollouts, excluding judge calls and superseded generation attempts.

\FloatBarrier
\section{Operating-Point Ablations}
\label{app:operating-point-ablations}

We detail the quality--throughput profiles in Section~\ref{sec:quality-throughput},
then report no-sharing baselines and controls for the recent window and query sharing.

\subsection{Quality and generation-cost sensitivity}
\label{app:adjusted-operating-points}

Figure~\ref{fig:quality-throughput} combines benchmark quality, controlled serving throughput, and generation length.
For profile $p$, let $A_{p,b}$ be benchmark $b$'s percentage score and $\bar L_{p,b}$ its mean generated tokens per response.
For $\mathcal B=\{\text{RULER 1M},\text{MRCR200},\text{GraphWalks-200},\text{AA-LCR}\}$, define
\begin{equation}
  A_p=\frac{1}{4}\sum_{b\in\mathcal B} A_{p,b},\qquad
  R_p=\frac{1}{4}\sum_{b\in\mathcal B}\frac{\bar L_{p,b}}{\bar L_{\mathrm{dense},b}},\qquad
  S_{\mathrm{adj},p}=\frac{S_p}{R_p},
  \label{eq:verbosity-adjusted-speedup}
\end{equation}
where $S_p$ is total-token throughput relative to the same model's dense reference.
Every profile includes all four benchmarks, weighted equally.
Output-length ratios are computed from each benchmark's mean, including reasoning tokens and outputs that reach the generation cap.
The base profile uses chunk budgets $\{4096,4096,4096,1024\}$ in the order above; the reduced profile halves each budget.
Each profile uses one sharing factor $G$ throughout, with a 128-page recent window and the current-page tail in addition to the selected chunks.

\paragraph{Evaluation subsets.}
Both models use identical tasks and reference answers.
\ruler{} uses its 1M-context evaluation, with 100 examples in each of 13 subtasks.
MRCR200 contains the longest 200 of the 2,400 MRCR prompts, selected before model-specific tokenization using source lengths of 763,613--1,118,929 tokens.
\graphwalks{}-200 contains the 200 prompts from Ultra's 256K--512K bucket, chosen to examine budget sensitivity; their framed lengths are approximately 360K tokens for Ultra and 255K for Inkling-Small.
MRCR200 and \graphwalks{}-200 each use four repeats per task; \aalcr{} uses 100 tasks with 16 repeats.
For each benchmark, quality and output length use the same tasks and repeats.

\paragraph{Throughput measurements.}
Each model serves the same ordered MRCR200 requests across its configurations, generating exactly 4,096 tokens per request with EOS ignored.
Ultra uses four B300 GPUs in TP2DP2 with concurrency 32 and temperature zero; each point averages three independent allocations with fresh servers.
The dense and base-$G=4$ runs share one campaign, while reduced-$G=4$ and base-$G=8$ use separate campaigns under the same workload and allocation protocol.
The $G=8$ campaign uses a newer serving build.
Inkling-Small uses eight B300 GPUs in TP8DP1, with concurrency and maximum sequence count 16, a maximum of 32,768 batched tokens, and one run per setting.
Throughput is the client-reported total-token rate from \texttt{vllm bench serve}, checked against token totals and elapsed time.
Speedups are normalized within each model; the different topologies do not support a controlled efficiency comparison between models.

\paragraph{Interpretation and uncertainty.}
Dividing by $R_p$ assumes cost is proportional to output length, omitting prefill cost, nonlinear batching effects, and workload differences.
The adjusted value is therefore a sensitivity proxy rather than measured natural-completion speedup.
Quality and throughput are measured separately, sometimes with different serving builds.
Ultra's dense \ruler{} reference also uses an earlier build and topology, with task and reference identities matched to the sparse runs.
Inkling-Small's quality and throughput runs use different builds.

Quality intervals use 20,000 task-bootstrap resamples, stratified within \ruler{}'s 13 subtasks and retaining all repeats of each sampled task for the other benchmarks.
Each resample recomputes the four-benchmark mean, using seed 20260901 for Ultra and 20260912 for Inkling-Small.
Intervals are pointwise, and Inkling-Small's single throughput run per setting provides no estimate of variation across runs.
Small differences between plotted points do not establish statistically resolved dominance.
Table~\ref{tab:quality-throughput-adjusted} reports both speedup coordinates and the quality intervals; Table~\ref{tab:quality-throughput-profile-components} gives Ultra's component scores.

\begin{figure}[t]
  \centering
  \includegraphics[width=0.98\textwidth]{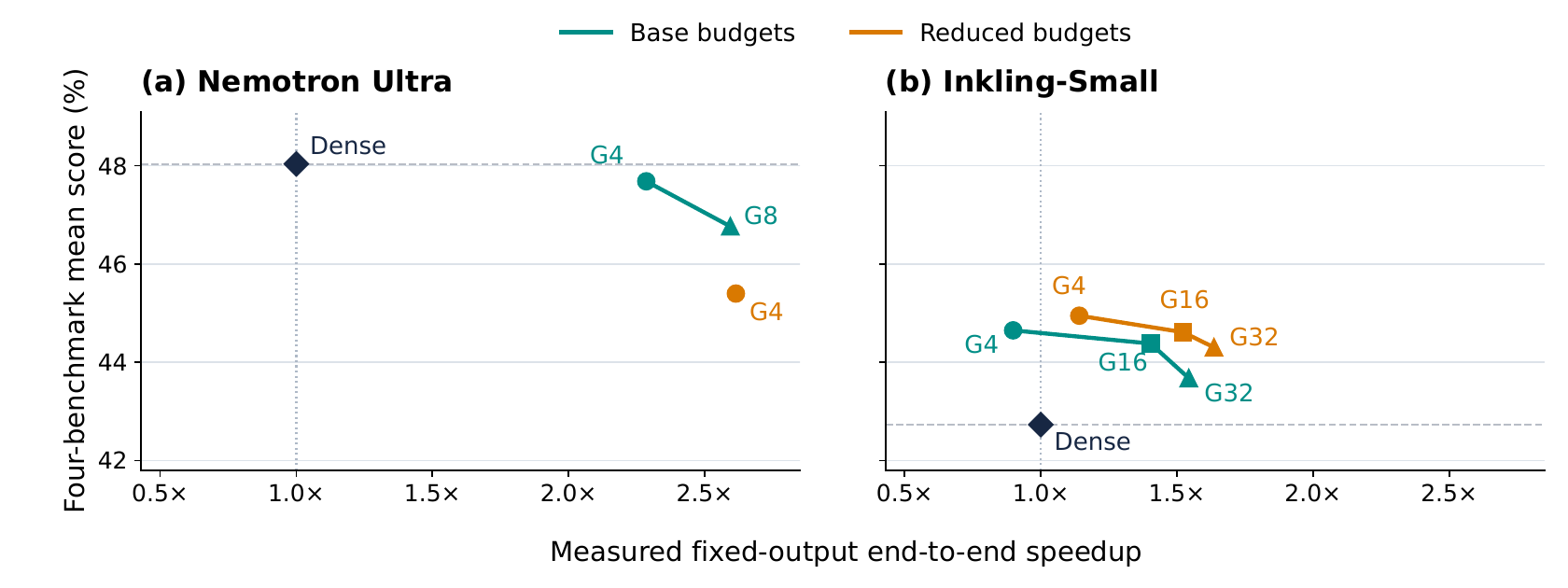}
  \caption{Unadjusted quality--throughput results for Ultra and Inkling-Small NVFP4.
  Quality scores and profiles match Figure~\ref{fig:quality-throughput}; the horizontal axis shows measured total-token throughput relative to each model's dense baseline, using MRCR200 with 4,096 output tokens per request.
  Lines connect increasing $G$ within each budget profile.
  Table~\ref{tab:quality-throughput-adjusted} reports the exact coordinates.}
  \label{fig:quality-throughput-unadjusted}
\end{figure}

\begin{table}[t]
  \centering
  \caption{Operating points for Figures~\ref{fig:quality-throughput} and~\ref{fig:quality-throughput-unadjusted}. Mean is the equally weighted four-benchmark score (\%) with its pointwise 95\% task-bootstrap interval. $S_p$ is measured fixed-output throughput relative to dense; $\Delta L=100(R_p-1)$ is the mean benchmark-relative output-length change. $S_{\mathrm{adj},p}=S_p/R_p$ is the generation-cost sensitivity proxy. Base budgets are $\{4096,4096,4096,1024\}$ chunks for RULER 1M, MRCR200, GraphWalks-200, and AA-LCR, respectively; reduced budgets halve these values.}
  \label{tab:quality-throughput-adjusted}
  \small
  \setlength{\tabcolsep}{3.5pt}
  \resizebox{\textwidth}{!}{%
  \begin{tabular}{lrrrrr}
    \toprule
    Profile & $G$ & Mean [95\% CI] & Measured $S_p$ & $\Delta L$ (\%) & Adjusted $S_{\mathrm{adj},p}$ \\
    \midrule
    \multicolumn{6}{l}{\textbf{Ultra NVFP4}} \\
    Dense & -- & 48.04 [45.75, 50.30] & $1.000\times$ & +0.0 & $1.000\times$ \\
    Base & 4 & 47.68 [45.39, 49.93] & $2.285\times$ & +17.9 & $1.939\times$ \\
    Reduced & 4 & 45.40 [43.14, 47.64] & $2.615\times$ & +34.1 & $1.950\times$ \\
    Base & 8 & 46.77 [44.48, 49.02] & $2.594\times$ & +13.9 & $2.277\times$ \\
    \midrule
    \multicolumn{6}{l}{\textbf{Inkling-Small NVFP4}} \\
    Dense & -- & 42.73 [40.34, 45.10] & $1.000\times$ & +0.0 & $1.000\times$ \\
    Reduced & 4 & 44.95 [42.65, 47.23] & $1.141\times$ & +75.1 & $0.652\times$ \\
    Reduced & 16 & 44.61 [42.28, 46.96] & $1.522\times$ & +76.6 & $0.862\times$ \\
    Reduced & 32 & 44.30 [41.93, 46.66] & $1.636\times$ & +76.6 & $0.926\times$ \\
    Base & 4 & 44.65 [42.34, 46.93] & $0.898\times$ & +39.7 & $0.643\times$ \\
    Base & 16 & 44.38 [42.04, 46.73] & $1.403\times$ & +38.6 & $1.013\times$ \\
    Base & 32 & 43.68 [41.32, 46.02] & $1.544\times$ & +37.3 & $1.124\times$ \\
    \bottomrule
  \end{tabular}}
\end{table}

\begin{table}[t]
  \centering
  \caption{Ultra NVFP4 benchmark scores underlying the operating-point means in Table~\ref{tab:quality-throughput-adjusted}.
  Scores use each benchmark's native percentage metric; the mean weights all four equally.}
  \label{tab:quality-throughput-profile-components}
  \small
  \setlength{\tabcolsep}{5pt}
  \begin{tabular}{lrrrrr}
    \toprule
    Profile & \ruler{} & MRCR200 & \graphwalks{}-200 & \aalcr{} & Mean \\
    \midrule
    Dense & 94.53 & 9.21 & 24.22 & 64.19 & 48.04 \\
    $G=4$, base & 93.76 & 9.99 & 22.43 & 64.56 & 47.68 \\
    $G=4$, reduced & 91.88 & 10.48 & 16.99 & 62.25 & 45.40 \\
    $G=8$, base & 92.69 & 9.81 & 20.69 & 63.88 & 46.77 \\
    \bottomrule
  \end{tabular}
\end{table}

\FloatBarrier
\subsection{No query sharing (\texorpdfstring{$G=1$}{G=1})}
\label{app:ultra-throughput-diagnostics}

To document performance without positional query sharing, Table~\ref{tab:no-query-sharing} reports Ultra NVFP4 quality and throughput separately.
The $G=1$ quality run uses $\chunkbudget=2{,}048$ and an implementation without a representative cache; its throughput run uses $\chunkbudget=4{,}096$.
These measurements describe different operating points and are not combined into one quality--throughput coordinate.

\begin{table}[!htbp]
  \centering
  \caption{Ultra NVFP4 results with and without query sharing.
  (a) \graphwalks{}-200 node-set F1 (\%) with pointwise 95\% intervals from 20,000 task-bootstrap resamples (seed 20260901).
  (b) MRCR200 total-token throughput on four B300 GPUs in TP2DP2 with concurrency 32 and 4,096 generated tokens per request.
  The $G=1$ throughput result is provisional and uses one allocation; other rows average three, with the observed range shown.}
  \label{tab:no-query-sharing}
  \small
  \begin{tabular}{lrrr}
    \toprule
    \multicolumn{4}{c}{\textbf{(a) Quality}} \\
    \midrule
    Setting & $k$ (chunks) & Score & 95\% CI \\
    \midrule
    Dense & -- & 24.22 & $[20.02,\,28.64]$ \\
    $G=1$ & 2,048 & 18.38 & $[14.67,\,22.30]$ \\
    $G=4$ & 2,048 & 16.99 & $[13.44,\,20.73]$ \\
    $G=4$ & 4,096 & 22.43 & $[18.32,\,26.73]$ \\
    $G=8$ & 4,096 & 20.69 & $[16.61,\,24.90]$ \\
    \bottomrule
  \end{tabular}

  \par\vspace{6pt}
  \begin{tabular}{lrrr}
    \toprule
    \multicolumn{4}{c}{\textbf{(b) Throughput}} \\
    \midrule
    Setting & $k$ (chunks) & Tokens/s & Range \\
    \midrule
    Dense & -- & 14,677 & 14,518--14,823 \\
    $G=1$ & 4,096 & 18,063 & -- \\
    $G=4$ & 2,048 & 38,375 & 38,083--38,767 \\
    $G=4$ & 4,096 & 33,543 & 33,091--34,196 \\
    $G=8$ & 4,096 & 38,075 & 37,656--38,434 \\
    \bottomrule
  \end{tabular}
\end{table}

Additional $G=1$ diagnostics give an \aalcr{} score of 65.4 over 15 of the planned 16 repeats and an MRCR200 score of 9.18 after recovering all saved responses; two MRCR200 servers failed their post-run health checks.
Earlier \ruler{} and MRCR200 runs from before an implementation repair are excluded.
Without a validated \ruler{} result and replicated throughput measurements, $G=1$ is excluded from the four-benchmark operating-point comparison.

\FloatBarrier
\subsection{Recent-window and query-sharing controls}
\label{app:routing-controls}

\paragraph{Matched recent-window control.}
On MRCR200 with four repeats per task, removing the 128-page recent window at $G=4,\chunkbudget=4{,}096$ changes the score from 9.99 to 9.64.
The paired difference is $-0.34$ percentage points before rounding, with a 95\% task-bootstrap interval of $[-1.44,0.71]$ (20,000 resamples, seed 20260901).
This experiment does not resolve an effect of the recent window on MRCR200 quality.

\begin{figure}[t]
  \centering
  \includegraphics[width=\textwidth]{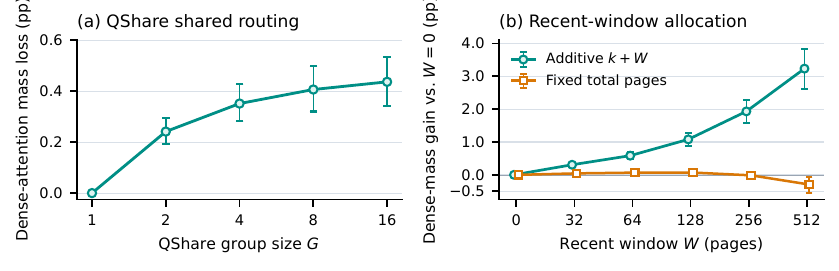}
  \caption{Offline routing diagnostics on 64 held-out 131K-token Ultra documents ($\chunkbudget=1{,}024$, 16-token pages). (a) Dense-attention-mass loss when one selection with $W=128$ is shared across $G$ adjacent queries over the final 2,048 positions. (b) Mass gain relative to $W=0$ when recent pages are added to $\chunkbudget$ selected pages or replace the same number at a fixed total, sampling within the final 8,192-position region. Whiskers are pointwise 95\% document-bootstrap intervals (2,000 draws, seed 42); window differences are paired with $W=0$. These are mechanism diagnostics, not benchmark-quality or throughput measurements.}
  \label{fig:routing-diagnostics}
\end{figure}

\paragraph{Held-out routing diagnostics.}
Figure~\ref{fig:routing-diagnostics} probes the two routing mechanisms directly on held-out activations from all 12 Ultra attention layers and both KV groups.
At $G=4$, \qshare{} retains, on average, 92.7\% of the pages selected independently at each position while losing 0.35 percentage points of total dense-attention probability mass.
The overlap between independent selections, measured as their intersection divided by the smaller set, is 88.3\% at a one-token lag and 87.4\% at lag 32.
For the recent window, adding $W=128$ pages to $\chunkbudget=1{,}024$ increases captured dense-attention mass by 1.08 percentage points, whereas replacing 128 selected historical pages at the same total page count adds 0.07 points.
In this diagnostic, most of the gain from adding the recent window is attributable to reading more pages.

\FloatBarrier
\section{Traffic Derivation}
\label{app:traffic-derivation}

We derive the traffic ratios in Section~\ref{sec:system}, omitting the recent window, current-page tail, and execution overheads.

For one KV group and one query with history $n$, dense attention reads $2n\headdim$ K/V elements.
After the sparse threshold $S=\chunkbudget\chunksize$, \cisa{} reads $2S\headdim$ selected K/V elements and $(n/\chunksize)\indexdim/\qsharegroup$ amortized representative elements.
Dividing by the dense read at $n=\ctxlen$ gives Equation~\ref{eq:traffic-ratio}.

For a complete causal prefill, dense K/V traffic is proportional to
$\int_0^{\ctxlen}2n\headdim\,dn=\ctxlen^2\headdim$.
\cisa{} attends densely through $S$ and then reads $2S\headdim$ selected K/V elements per query, giving
$S^2\headdim+2S(\ctxlen-S)\headdim$.
The representative scan is active only over the indexed-history region and integrates to
$(\ctxlen^2-S^2)\indexdim/(2\chunksize\qsharegroup)$.
Normalizing by dense traffic and substituting $\alpha=S/\ctxlen$ gives Equation~\ref{eq:prefill-traffic}.
An additive recent window and current-page tail introuduce additional traffic that is omitted from these  expressions. The inverse ratios are therefore bandwidth-only estimates, not strict bounds on measured speedup.

The representative cache has the following capacity relative to the main K+V state:
\begin{equation}
  \frac{(\ctxlen/\chunksize)\indexdim}{2\ctxlen\headdim}
  =\frac{\indexdim}{2\chunksize\headdim}.
\end{equation}
For $\chunksize=16$ and $\indexdim=\headdim$, this is $1/32$.
This capacity is local to the KV group and follows its tensor-parallel ownership.

\section{Attention-Only Speedup Measurements}
\label{app:attention-only-speedup}

These one-layer experiments measure the attention-path costs discussed in Section~\ref{sec:component-to-throughput} for Nano, Ultra, and Inkling-Small.

\subsection{Nano and Ultra head geometries}
\label{app:nemotron-attention}

We compare dense and sparse one-layer Nemotron-H models with dummy weights on one B300 GPU at TP1.
All runs use CUDA graphs, BF16 model state and K/V, head dimension 128, 32 query heads, and a maximum of 8,192 batched tokens.
Two KV heads reproduce Nano's TP1 attention geometry; one KV head with 32 query heads reproduces Ultra's local geometry at TP2, excluding inter-GPU communication.
\cisa{} uses 16-token chunks, $\chunkbudget=4{,}096$, a 128-page recent window, and $G=4$ for prefill; single-token decode has effective $G=1$.

For batch sizes $B\in\{16,24,32\}$, we retain ten terminal-prefill or steady-decode forwards near $N=1{,}048{,}576$.
Profiler annotations verify the phase, batch size, and per-request history.
Table~\ref{tab:attention-only-speedup} reports speedups over dense GQA for each batch size and their geometric mean.

\begin{table}[t]
  \centering
  \caption{\cisa{} speedup over matched dense GQA at one million tokens (higher is better).
  The geometric mean is across the three executed batch sizes.
  GPU busy is the union of kernel, memcpy, and memset intervals, with overlaps counted once.
  Core attention includes only the FlashAttention forward and combine kernels; GPU busy also includes projections, scoring, selection, metadata, and copies.}
  \label{tab:attention-only-speedup}
  \scriptsize
  \setlength{\tabcolsep}{4pt}
  \resizebox{\textwidth}{!}{%
  \begin{tabular}{lllrrrr}
    \toprule
    Local model shape & Phase & Metric & $B=16$ & $B=24$ & $B=32$ & Geom. mean \\
    \midrule
    \multirow{4}{*}{Nano TP1 ($H_{\mathrm{KV}}=2$)}
      & \multirow{2}{*}{Terminal prefill} & GPU busy & $4.49\times$ & $4.57\times$ & $4.56\times$ & $4.54\times$ \\
      & & Core attention & $5.95\times$ & $6.40\times$ & $6.07\times$ & $6.13\times$ \\
      \cmidrule(lr){2-7}
      & \multirow{2}{*}{Steady decode} & GPU busy & $2.72\times$ & $3.24\times$ & $3.39\times$ & $3.11\times$ \\
      & & Core attention & $12.37\times$ & $13.18\times$ & $12.94\times$ & $12.83\times$ \\
    \midrule
    \multirow{4}{*}{Ultra TP2/rank ($H_{\mathrm{KV}}=1$)}
      & \multirow{2}{*}{Terminal prefill} & GPU busy & $8.61\times$ & $8.09\times$ & $8.61\times$ & $8.43\times$ \\
      & & Core attention & $11.63\times$ & $11.57\times$ & $11.70\times$ & $11.63\times$ \\
      \cmidrule(lr){2-7}
      & \multirow{2}{*}{Steady decode} & GPU busy & $2.56\times$ & $2.96\times$ & $2.98\times$ & $2.83\times$ \\
      & & Core attention & $13.26\times$ & $13.79\times$ & $12.99\times$ & $13.34\times$ \\
    \bottomrule
  \end{tabular}}
\end{table}

At this context length, the selected K/V span is $\chunkbudget\chunksize=65{,}536=N/16$.
The ideal K/V-only reduction is therefore $16\times$, or $14.22\times$ after accounting for the representative scan at $G=4$ (Equation~\ref{eq:traffic-ratio}).
Ultra's geometric-mean terminal-prefill speedup is $11.63\times$ for core attention and $8.43\times$ for all GPU activity.
The remaining costs include the recent window, current-page tail, routing, metadata, projections, and copies.
Nano's lower prefill speedup reflects an implementation that processes its two local KV groups serially on one CUDA stream, while dense attention uses a single kernel.

During decode, the representative scan cannot be shared across positions, giving an ideal combined traffic reduction of $10.67\times$ at $G=1$.
Core-attention speedups are $12.83\times$ for Nano and $13.34\times$ for Ultra, but including all GPU activity reduces them to $3.11\times$ and $2.83\times$.
These gaps illustrate the cost of work outside the core attention kernel.

At $B=24$, the 8,192-token batching cap gives 341 prefill tokens per request and a final one-token tail.
Dense and sparse runs execute the same shapes, including this irregular tail.

\clearpage
\subsection{Inkling-Small component breakdown}
\label{app:inkling-attention}

Table~\ref{tab:inkling-attention-breakdown} gives the timings behind Figure~\ref{fig:inkling-systems-breakdown}, including the decode implementation before coalescing representative loads.
The prefill sweep varies query sharing at a fixed chunk budget; the decode optimization preserves the scoring rule and selection policy.

\paragraph{Model and hardware.}
We use one B300 GPU at TP1 with a one-layer dummy-weight model containing global attention, a dense MLP, and short convolution.
Four query heads and one KV head reproduce Inkling-Small's local attention geometry at TP8.
Hidden and MLP intermediate dimensions are 4,096, convolution width is 4, and attention and representative dimensions are 128.
The model omits MoE, sliding-window layers, and TP8 communication.
Model state, activations, representatives, and K/V are BF16; execution uses FlashAttention~4, CUDA graphs, and the baseline top-$k$ implementation.

\paragraph{Workload and measurements.}
The workload comprises sixteen 1,048,576-token prompts, with a maximum of 8,192 batched tokens and a 64\,GiB KV reservation.
Sparse runs use 16-token chunks, $\chunkbudget=4{,}096$, a 128-page recent window, and the current-page tail.
Prefill varies $G\in\{4,16,32\}$; decode uses the $G=32$ configuration with one query per sequence, giving effective positional sharing $G=1$.
Each mean uses ten forwards from one capture, rather than independent trials.
Terminal-prefill forwards contain 512 queries per sequence at histories 1,043,968--1,048,576 in steps of 512; decode histories span 1,048,694--1,048,703.
The baseline and coalesced-load decode implementations are measured on the same B300; the dense decode reference is captured on another B300.

\begin{table}[!htbp]
  \centering
  \caption{Inkling-Small attention-path timings for Figure~\ref{fig:inkling-systems-breakdown}. Values are mean milliseconds over ten forwards from one capture. GPU busy counts overlapping kernel and memory-operation intervals once and excludes idle gaps. Component sums may differ from totals because of rounding.}
  \label{tab:inkling-attention-breakdown}
  \scriptsize
  \setlength{\tabcolsep}{3pt}
  \resizebox{\textwidth}{!}{%
  \begin{tabular}{llrrrrrr}
    \toprule
    Phase & Variant & Attention & Scoring & Top-$k$ & Setup & Other & GPU busy \\
    \midrule
    Terminal prefill & Dense & 14.176 & 0.000 & 0.000 & 0.018 & 0.912 & 15.106 \\
    Terminal prefill & CISA G4 & 9.154 & 1.174 & 1.004 & 0.152 & 0.972 & 12.456 \\
    Terminal prefill & CISA G16 & 2.626 & 0.589 & 0.325 & 0.069 & 0.968 & 4.577 \\
    Terminal prefill & CISA G32 & 1.198 & 0.587 & 0.147 & 0.060 & 0.962 & 2.954 \\
    \midrule
    Steady decode & Dense & 1.218 & 0.000 & 0.000 & 0.019 & 0.092 & 1.329 \\
    Steady decode & CISA & 0.104 & 0.536 & 0.118 & 0.047 & 0.112 & 0.918 \\
    Steady decode & CISA + coalesced loads & 0.104 & 0.057 & 0.118 & 0.045 & 0.111 & 0.436 \\
    \bottomrule
  \end{tabular}}
  \vspace{4pt}
  \begin{minipage}{\textwidth}
  \footnotesize
  \textbf{Components.} Attention contains core attention kernels; Scoring contains representative scoring; Top-$k$ contains selection. Setup includes scorer-plan construction, query sharing, representative maintenance, page construction, and block-table gather. Other contains the remaining kernels and copies, with overlaps counted once.\par
  \end{minipage}
\end{table}

\end{document}